%% file: paper.tex
\documentclass{amia}
\usepackage{graphicx}
\usepackage[labelfont=bf]{caption}
\usepackage[superscript,nomove]{cite}
\usepackage{color}
\usepackage{commands}
\usepackage{amsmath}
\usepackage{amsfonts}
\usepackage{amssymb}
\usepackage{tabularx}
\usepackage{tikz}
\usepackage{url}
\begin{document}

\title{Named Entity Recognition for Electronic Health Records: A Comparison of Rule-based and Machine Learning Approaches}

\author{Philip John Gorinski$^1$, Honghan Wu$^{2}$, Claire Grover$^{1}$, Richard Tobin$^{1}$, Conn Talbot$^{3}$, Heather Whalley$^{3}$, Cathie Sudlow$^{2}$, William Whiteley$^{3}$, Beatrice Alex$^{1,4}$}

\institutes{
    $^1$Institute for Language, Cognition and Computation, School of Informatics, University of Edinburgh; $^2$Usher Institute, University of Edinburgh; $^3$Centre for Clinical Brain Sciences, University of Edinburgh; $^4$Edinburgh Futures Institute, University of Edinburgh\\
}

\maketitle

\noindent{\bf Abstract}

\textit{This work investigates multiple approaches to Named Entity Recognition (NER) for text in Electronic Health Record (EHR) data. In particular, we look into the application of (i) rule-based, (ii) deep learning and (iii) transfer learning systems for the task of NER on brain imaging reports with a focus on records from patients with stroke. We explore the strengths and weaknesses of each approach, develop rules and train on a common dataset, and evaluate each system's performance on common test sets of Scottish radiology reports from two sources (brain imaging reports in ESS -- Edinburgh Stroke Study data collected by NHS Lothian as well as radiology reports created in NHS Tayside). Our comparison shows that a hand-crafted system is the most accurate way to automatically label EHR, but machine learning approaches can provide a feasible alternative where resources for a manual system are not readily available.}

\section*{Introduction}
Named Entity Recognition (NER) is an area of Natural Language Processing (NLP) that addresses the identification and classification of entities in written text. It has been employed using large variety of methods and on a multitude of domains and methods.\cite{Nadeau2007,Leaman2008,Ritter2011,Rocktaschel2012}

\begin{figure}[h]
\frame{
	\includegraphics[width=\textwidth]{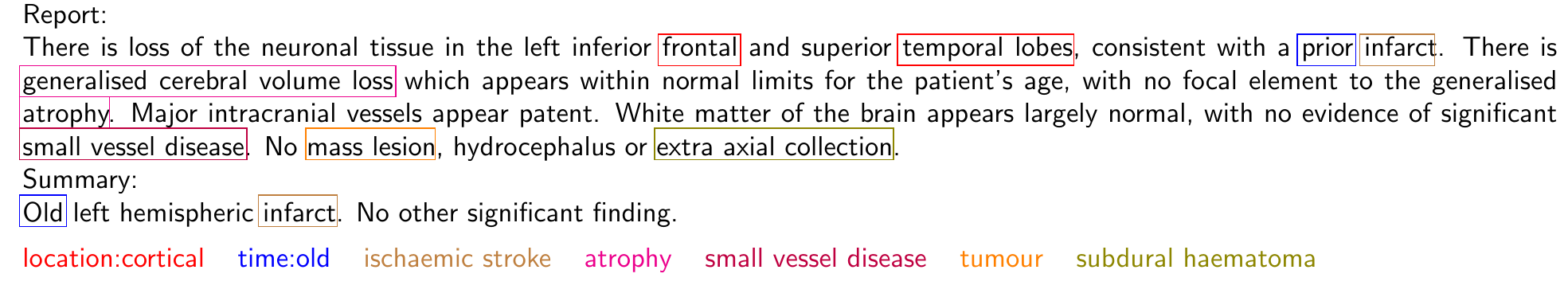}}
	\vspace{0.3cm}\caption{Example of a brain imaging report with annotated entities and their types below.}
	\label{fig:ehr}
\end{figure}

Electronic Health Records (EHR) typically contain not only structured information about a patient but also written, unstructured text describing health professionals' opinions.  Named entities in this domain include names of diseases, symptoms and anatomical locations. A radiology report is the opinion of a radiologist on a scan or X-ray.  Figure~\ref{fig:ehr} shows an example of an anonymised radiology report of a brain scan with identified named entities.

\section*{Related Work}

NER is a well-studied field of NLP.\cite{McCallum2003,Nadeau2007,Ratinov2009} In 2003, Tjong and et al.\cite{tjong2003} introduced a shared NER task at the Conference on Computational Natural Language Learning (CoNLL), which established a widely-accepted benchmark for the evaluation of NER systems. This led to research into machine learning methods, such as the Stanford NER tagger.\cite{finkel2005} Other NER systems follow a rule-based approach, such as the ANNIE NER tagger.\cite{cunningham2002}

NLP for the medical domain has been an active field of research since the early 2000's. BioCreative and BioNLP provided shared tasks for NER and Relation Extraction (RE), with several systems applying NLP to biomedical text.\cite{Settles2004,Leaman2008,Ogren2007,Savova2010,alex2007,grover2007} An overview of approaches to information extraction from EHR data was conducted by Meystre et al. (2008)\cite{meystre2008}, and Pons et al. (2016)\cite{pons2016} provide a recent review of NLP in radiology.

Most relevant for the systems investigated in the current work in terms of domain is work conducted by Flynn et al. (2010)\cite{flynn2010}, who present a system for the analysis of brain scan radiology reports. While not dealing with NER as a their main task, the authors applied keyword matching to analyse reports from the Tayside dataset, and assign document-level labels differentiating between stroke type (ischaemic stroke versus intracerebral haemorrhage).

There are many Machine Learning/Deep Learning architectures proposed in the literature for NER. In this paper, we draw from the line of work presented in Huang at al. (2015)\cite{Huang2015}, who employed Conditional Random Fields (CRF)\cite{Lafferty2001} on top of bidirectional Long Short-Term Memories (LSTM).\cite{schuster1997}  Cornegruta et al. (2016)\cite{Cornegruta2016} evaluated a NER method on radiology reports.  They employed a bidirectional LSTM (BiLSTM) neural network architecture, which they contrasted with a simple baseline of string matching against external lexicons.

% might be a good idea to include transfer learning in this domain
Transfer learning\cite{Pan2010} methods reuse machine learning models originally trained for a source task in a new target task. This idea has been adopted for doing NER tasks in a transferable manner, e.g.~Arnold et al (2008)\cite{Arnold2008} used feature hierarchy, Nothman et al. (2013)\cite{Nothman2013} utilised the text and structure of Wikipedia, and Collobert and others created a convolutional neural network to jointly train multiple tasks.\cite{Collobert2008}

\section*{Data}
The datasets we used to perform NER consist of anonymised radiology reports from brain MRI and CT scans conducted as part of the Edinburgh Stroke Study (ESS)\cite{Jackson2008} (n=1,168) and routine scans conducted by NHS Tayside (n=156,619).  From the ESS data, a subset of 630 reports were annotated. From the Tayside collection, two subsets (Tay and TayExt) were selected and annotated.  Each subset consists of reports for training/development of NER systems, as well as held-out test reports for testing and system comparison (see Table~\ref{tab:data} for some statistics). 

\begin{table}[h]
	\small
	\begin{center}
		\begin{tabular}{l|rr|rr|rr}
			& ESS dev & ESS test & Tay dev & Tay test & TayExt dev & TayExt test\\
			\hline
			\#Reports & 364 & 266 & 362 & 700 & 1,068 & 300\\
			\#Sentences & 3,837 & 2,855 & 2,791 & 3,948 & 8,401 & 2,677\\
			% ESS sentence counts updated to include all sentences (not just ones with a proc=yes attribute), so counting all sentence with an id attribute (s[@id])
			%\#Sentences & 3,731 & 2,755 & 2,791 & 3,948 & 8,401 & 2,677\\
			\#Named Entities & 4,332 & 2,924 & 2,997 & 2,986 & 7,642 & 2,637
			%Tayside test counts were fixed, after realising one count in each devtest and test was an annotation error (an entity annotation on a document level label)
			%\#Named Entities & 4,332 & 2,924 & 2,998 & 2,987 & 7,642 & 2,637
		\end{tabular}
	\end{center}
	\caption{Number of reports, sentences and named entities per subset and development/training (dev) and test splits.}
	\label{tab:data}
\end{table}

ESS was the first set to be annotated by domain experts, and the rule-based EdIE-R system\cite{Alex2019} was developed on this dataset. Data from NHS Tayside (Tay) was subsequently annotated with the same annotation scheme. This not only provides us with additional data, but also introduces different distributions of entities.  This difference in data was further amplified by a second round of annotation on Tayside (TayExt), with reports specifically selected to include low-frequency entities.\footnote{TayExt is a subset of Tayside brain imaging reports which mention one of a list of keywords (e.g. bleed*, subarachnoid, subdural, haemorrh*, hemorrh*, mass, tumour or tumor).   This filtering was done to ensure that certain entities which appeared infrequently in the previous datasets would be more frequent.}  Detailed frequency counts for entities annotated in ESS, Tay and TayExt are shown in Table~\ref{tab:ent-freq}.

\begin{table}[h]
	\small
	\begin{center}
		\begin{tabular}{lrrrrrr}
			Entity Type & ESS dev & ESS test & Tay dev & Tay test & TayExt dev & TayExt test\\
			\hline
			ischaemic stroke & 697 & 455 & 369 & 306 & 668 & 214\\
			haemorrhagic stroke & 344 & 267 & 428 & 294 & 890 & 280\\
			stroke & 60 & 26 & 32 & 9 & 33 & 5\\
			glioma tumour & 0 & 0 & 10 & 9 & 32 & 12\\
			meningioma tumour & 4 & 8 & 9 & 2 & 32 & 6\\
			%metastasis tumour & 24 & 12 & 61 & 120 & 117 & 35\\
			metastasis tumour & 24 & 12 & 61 & 119 & 117 & 35\\
			tumour & 297 & 166 & 146 & 303 & 432 &117\\
			subdural haematoma & 244 & 109 & 75 & 95 & 968 & 309\\
			small vessel disease & 427 & 276 & 61 & 173 & 288 & 74\\
			atrophy & 246 & 153 & 105 & 168 & 350 & 90\\
			microhaemorrhage & 12 & 10 & 0 & 6 & 1 & 2\\
			subarachnoid haemorrhage & 13 & 10 & 49 & 16 & 135 & 54\\
			%subarachnoid haemorrhage & 13 & 10 & 50 & 16 & 135 & 54\\
			haemorrhagic transformation & 5 & 2 & 16 & 1 & 44 & 10\\
			location:cortical & 516 & 412 & 924 & 476 & 1775 & 665\\
			location:deep & 524 & 343 & 299 & 574 & 697 & 273\\
			time:old & 527 & 321 & 250 & 158 & 558 & 218\\
			time:recent & 392 & 354 & 163 & 277 & 622 & 273\\
			%Tayside test counts were fixed, after realising one count in each devtest and test was an annotation error (an entity annotation on a document level label)
		\end{tabular}
	\end{center}
	\caption{Per-entity frequency counts in the ESS, Tay and TayExt development (dev) and test datasets.}
	\label{tab:ent-freq}
\end{table}

Each set contains rich annotations of named entities in the text but also includes negated entities, entity relations and document-level labels.  In this paper, we only focus on the entity annotation, not distinguishing between positive and negative entities. To ensure consistency, a first round of annotations from different annotators were compared before annotators carried out their work for the full datasets. Annotators showed very high inter-annotator agreement (IAA) on the test data sets (see \emph{IAA} column in Tables~\ref{tab:exp-ess-results} and~\ref{tab:exp-tay-results} in the Experiments section).  We report IAA figures for the entire ESS test data and for a subset of 100 reports from the Tayside test data.\footnote{We do not report results for the TayExt data because double annotation for that dataset has yet to be carried out.}

\section*{NER System Descriptions}

For our comparative experiments on NER performance, we chose to evaluate a rule-based, a deep learning and a transfer learning system which we introduce here.

\subsection*{EdIE-R}
\begin{figure}[h]
	\begin{center}
		\includegraphics[width=.55\textwidth]{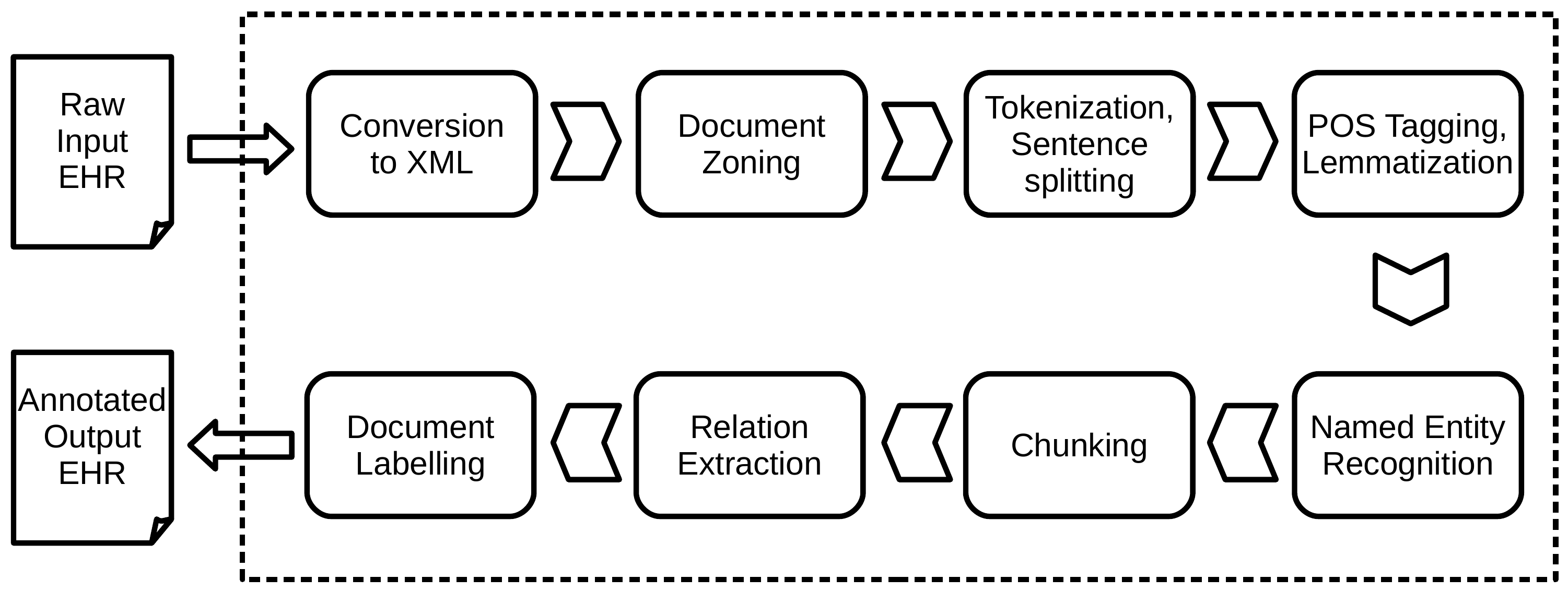}
	\end{center}
	\caption{Overview of the \emph{EdIE-R} pipeline.}
	\label{fig:edie-r}
\end{figure}

\emph{EdIE-R} (Edinburgh Information Extraction for Radiology reports)\cite{Alex2019} is a rule-based system. It consists of a full pipeline that starts with the raw input text, and subsequently adds sectioning, tokenisation, sentence-splitting and linguistic annotation such as part-of-speech (POS) tagging and shallow syntactic analysis.\footnote{The tokenised and POS tagged output of EdIE-R was used to prepare the datasets used for evaluation by all systems described in this paper.}  Figure~\ref{fig:edie-r} provides a schematic overview of the \emph{EdIE-R} pipeline.

Of particular interest to the comparative evaluation presented in this work is the NER step of \emph{EdIE-R}. At this stage in the pipeline, the raw text has already been tokenised and POS tagged. Making use of hand-crafted rules and lexicons created in consultation with radiology experts, the system then uses the information derived during the previous steps to perform NER for specific target entities.  

\emph{EdIE-R} has been shown to recognise named entities reliably accurately in brain imaging reports in the ESS data, which was used to write the original NER rules for this domain.\cite{Alex2019} We have subsequently updated the rules based on the new development data from Tayside and all the results reported here are from the updated version. Performance on the ESS data has dropped very slightly from the earlier version but \emph{EdIE-R} performs very well on the new data. The reliance on hand-crafted rules makes it potentially costly and time-consuming to adapt the system to a different dataset, for example, radiology reports for scans of other body parts or for other diseases as well as other types of raw text records such as pathology reports.

The initial \emph{EdIE-R} rule writing was done iteratively in parallel with rounds of annotation done by domain experts before settling on an annotation scheme.  Several rounds of annotation were carried out to create gold data for system development and evaluation (ESS, Tay and TayExt).  Having this annotated gold data available provided us with the opportunity to try and test machine learning based methods which are typically used for NER on standard evaluation datasets (e.g. CoNLL or ACE data).

\subsection*{EdIE-N}
\emph{EdIE-N} represents a machine learning based approach to the problem of NER for radiology reports.  As opposed to \emph{EdIE-R}'s hand-crafted rules, \emph{EdIE-N} infers named entity annotation from training data automatically, and applies these learned ``rules'' captured by the trained model to new data.

\begin{figure}[h]
	\begin{center}
		\includegraphics[width=.58\textwidth]{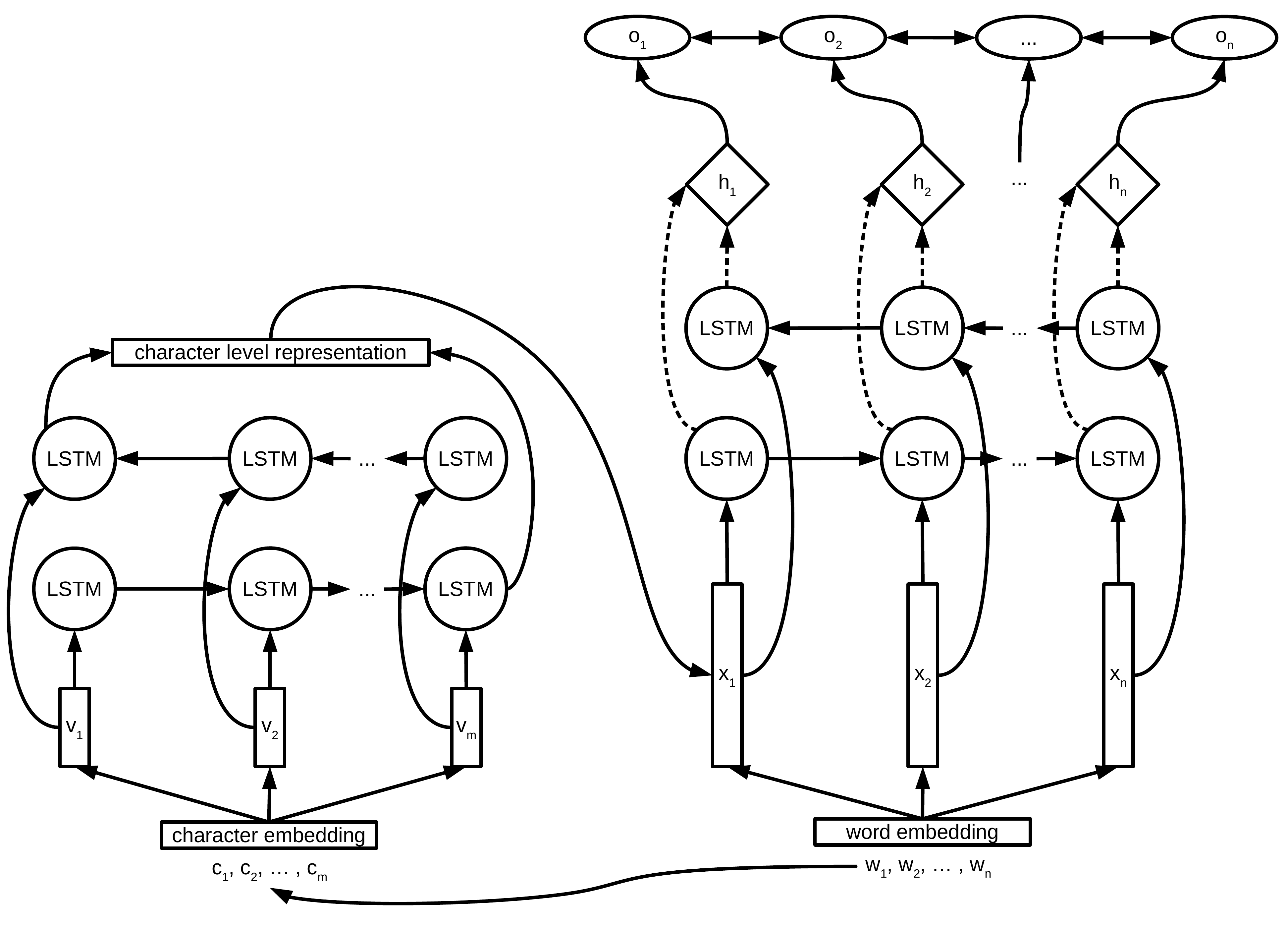}
	\end{center}
	\caption{Schematic of \emph{EdIE-N} entity recognition network.}
	\label{fig:edie-n}
\end{figure}

In particular, the system makes use of \emph{deep learning} via a neural network architecture (see Figure~\ref{fig:edie-n}). \emph{EdIE-N} employs a Conditional Random Field (CRF)\cite{Lafferty2001} on top of a bi-directional LSTM \cite{schuster1997} architecture to perform NER by assigning the best score $s$ of consecutive of output labels $o$ to a given sentence. The input features $x_{i}$ of the classification network are comprised of word embeddings, which get concatenated with word representations derived from a character-level LSTM. This architecture is similar that of Huang et al. (2015)\cite{Huang2015} and Zhou et al. (2015)\cite{Zhou2015} (see Figure~\ref{fig:edie-n}).

Both word embeddings as well as character embeddings can either be learned during training from randomly initialized embedding matrices, or looked up in pre-trained models.  At training time, the CRF output layer is conditioned on the LSTM hidden layer representations $h_{i}$. At test time, the system assigns entity annotations to each word according to the most likely entity type as determined by the CRF.

EdIE-N models can be trained either as a ``monolithic'' NER model, i.e.~taking all possible entity types into account and potentially making use of interactions between them, or as a ``bag-of-models" system, where one model is trained per unique entity type. We only report the results of the ``bag-of-models'' setup as using this approach makes it easier to add new entity types to an existing architecture.  However, we have experimented with ``monolythic'' NER models which resulted in broadly similar performances to the latter.

As opposed to \emph{EdIE-R}, the machine learning approach employed by \emph{EdIE-N} does not rely on hand-crafted rules for NER. Instead, the system is trained on an annotated gold standard, from which entity type assignments are learned automatically. This alleviates the need for expert knowledge for designing new rules, making it both fast and inexpensive to learn and abstract from any given dataset. However, as a fully supervised machine learning approach, it does introduce the need for annotated gold data for training.  Moreover, there is a common understanding that a sufficiently large training dataset is needed for a model to learn enough examples so that it performs reasonably well on new data.  Creating such data is time-consuming.  The other limitation to a machine learning based system is that it is very difficult to conduct error analysis and determine the exact reason for system errors.

\subsection*{SemEHR}
\input{semehr.tex}

\section*{Experiments}

We used strict CoNLL-style NER evaluation to compare the performance of the three systems on the different datasets and report individual scores per entity type and overall NER scores.  We report precision (P), recall (R) and balanced F-score (F1), the harmonic mean of precision and recall.  In the case of EdIE-N we average scores over 5 runs to account for fluctuations in classification results due to random initializations in the network models.

For EdIE-N first we report overall scores when training is performed on the development data of ESS, ESS plus Tay and all three development sets combined (ESS+Tay+TayExt).  While the model trained on ESS data performs best on its own test data, we consider EdIE-N trained on all three datasets to be the better one as it performs best on the other two test sets and only slightly worse on ESS test (see Table~\ref{tab:edie-n-models}). We use this model for the subsequent comparison.

\begin{table}[h]
	\small
	\begin{center}
		\begin{tabular}{l|ccc|ccc|ccc}
			Evaluation on Test data & \multicolumn{3}{c|}{ESS} &  \multicolumn{3}{c|}{Tay}  &  \multicolumn{3}{c}{TayExt} \\\hline
			Training data & P & R & F1 & P & R & F1 & P & R & F1\\
			\hline
			ESS & \textbf{0.85} & \textbf{0.93} & \textbf{0.89} & 0.75 & 0.83 & 0.79 & 0.69 & 0.81 & 0.75 \\
			ESS+Tay & 0.84 & 0.91 & 0.87 & 0.78 & 0.87 & 0.83 & 0.75 & 0.83 & 0.79 \\
			ESS+Tay+TayExt & 0.82 & 0.92 & 0.86 & \textbf{0.80} & \textbf{0.91}& \textbf{0.85} & \textbf{0.76} & \textbf{0.85} & \textbf{0.80} \\
		\end{tabular}
	\end{center}
	\caption{EdIE-N performance under different combinations of training and test data (best scores in \textbf{boldface}).}
	\label{tab:edie-n-models}
\end{table}

\input{expresults.tex}

In our final experiment, we compare the rule-based \emph{EdIE-R} system against the \emph{EdIE-N} LSTM-CRF architecture and the \emph{SemEHR} transfer learning approach on the ESS, Tayside, and extended Tayside test sets (see Tables~\ref{tab:exp-ess-results}, \ref{tab:exp-tay-results} and \ref{tab:exp-tayext-results}).

The rule-based system \emph{EdIE-R} outperforms both machine learning approaches, even reaching near IAA levels on the ESS data. The gap between \emph{EdIE-R} and \emph{SemEHR}, the transfer learning approach incorporating rich information from out-of-domain resources, is relatively small (ranging between 0.03 and 0.06 points in F1), especially on the ESS and extended Tayside data.  It is a little more pronounced on the original Tayside test set. \emph{EdIE-R} clearly benefits from tailored domain and data specific rules. Overall, \emph{SemEHR} performs remarkably accurately on the test data, matching the hand-crafted system for a few of the entity types. On all datasets, the machine learning approach applied by \emph{EdIE-N} falls behind the two other systems. However, these results are of little surprise, as \emph{EdIE-N} uses no external knowledge such as access to an ontology, and relies entirely on features that are being derived automatically from the target texts. It is very likely that the performance of \emph{EdIE-N} can be further improved by using additional training data, incorporating additional domain knowledge, or optimising model parameters further. While \emph{EdIE-R} is the best overall system, there are certain labels, e.g.~\emph{subarachnoid haemorrhage}, for which \emph{SemEHR} consistently performs better.

\section*{Conclusions}
We have presented a system comparison for the task of labelling Named Entities in Electronic Health Records. Three approaches to the task were evaluated on three data sets. A hand-written system engineered by domain experts was able to consistently outperform a transfer learning system, applying previously established rules to new data, and a data-driven machine learning system.

The results confirm previously established findings that a hand-written, rule-based approach is able to perform NER on written EHR data very accurately, albeit for a high development cost in terms of time and effort afforded by domain experts. While the machine learning approach performed worse in our comparison, we are still slightly optimistic that such an approach can be reasonably employed where there are either no experts readily available, or a system has to be developed quickly for a relatively low cost. The transfer learning approach showed impressive results, presenting a viable alternative to an entirely hand-written system, though still requiring a good deal of human manipulation.

In the future, we would like to further improve the ways to reliably and automatically label Named Entities in EHRs. Of particular interest are more experiments on the machine learning approach, where especially more fine-grained tuning of hyper parameters promises to be able to yield better performance results. Additionally, we would like to explore the possibility of combining the approaches presented in this paper. The modular nature of the overall EHR processing pipeline could enable us to employ the different NER systems according to their individual strengths and weaknesses, potentially leading to a better overall performance downstream, e.g, at the document labelling stage. Another interesting future direction is the rapid development of new systems in multiple iterations. When required, one could start by rapidly and inexpensively adding a new label to the system using the machine learning approach, and subsequently improving on it by utilising its results to guide the transfer or hand-crafting of reliable rules.

\section*{Acknowledgements}
Gorinski, Tobin, Grover, Alex and Whalley are supported by the MRC Mental Health Data Pathfinder Award (MRC - MC\_PC\_17209). Wu is MRC/Rutherford fellow of HRD UK (MR/S004149/1). Grover was and Alex is supported by The Alan Turing Institute (EPSRC grant EP/N510129/1). Whiteley was supported by an MRC Clinician Scientist Award (G0902303) and is supported by a Scottish Senior Clinical Fellowship (CAF/17/01). Sudlow is Chief Scientist of UK Biobank and Director of HDR UK Scotland. 

\makeatletter
\renewcommand{\@biblabel}[1]{\hfill #1.}
\makeatother

\bibliographystyle{unsrt}

\end{document}

%% file: semehr.tex
The third approach we chose to compare to is a NER tool which was originally developed and trained for other purposes. The main goal is to compare the above two approaches with a generic portable tool that is able to be adapted for this particular stroke subtyping task. The tool picked for this purpose is \emph{SemEHR}~\cite{wu2018semehr}, which is an open source toolkit that integrates text mining and semantic computing for identifying mentions of UMLS~\cite{umls2004} (Unified Medical Language System) concepts from clinical documents. Specifically, we adopted a \emph{SemEHR} instance that has been trained on EHR data of South London and Maudsley, a psychiatric hospital in London. This instance was trained for identifying physical illnesses, such as liver diseases, HIV, diabetes etc. Each mention identified by \emph{SemEHR} was associated with three-dimensional contextual information, i.e.~negation (whether the condition was negated or affirmed), temporality (whether it was a recent or past event) and experiencer (whether the sufferer was the patient or other people).

\emph{SemEHR} is based on GATE Bio-YODIE~\footnote{https://gate.ac.uk/applications/bio-yodie.html} and was adapted in two steps. The first step involved generating a mapping from what \emph{SemEHR} identifies (i.e.~mentions of UMLS concepts) to what this study is looking for (i.e.~the entity types listed in Table~\ref{tab:ent-freq}). For those entity types not in UMLS vocabulary (e.g.~\emph{small vessel disease}), an additional dictionary is generated and combined with \emph{SemEHR}'s existing gazetteer.  There are cases where one UMLS concept is mapped to different entity types (e.g.~\emph{C0038454} is mapped to \emph{stroke} and \emph{ischaemic stroke}). To disambiguate them, the second adaptation step was to train a machine learning model on those cases. Details and source code of the second step are made available on GitHub.\cite{semEHRcode}

%https://github.com/CogStack/nlp2phenome

%% file: expresults.tex
\begin{table}[h!]
	\small
	\begin{center}
		\begin{tabular}{l|rrr|rrr|rrr||rrr}
			Evaluation on ESS test & \multicolumn{3}{c|}{EdIE-R} & \multicolumn{3}{c|}{EdIE-N}  & \multicolumn{3}{c}{SemEHR} & \multicolumn{3}{c}{IAA} \\
			\hline
			Entity Type  & P & R & F1 & P & R & F1 & P & R & F1 & P & R & F1\\
			 \hline
			ischaemic stroke & \textbf{1.00} & \textbf{1.00} & \textbf{1.00} & 0.87 & 0.96 & 0.92 & 0.96 & \textbf{1.00} & 0.98 & 0.98 & 1.00 & 0.99\\
			haemorrhagic stroke & 0.80 & 0.84 & 0.82 & 0.74 & 0.76 & 0.75 & \textbf{0.86} & \textbf{0.93} & \textbf{0.90} & 0.93 & 0.99 & 0.96\\
			stroke & \textbf{1.00} & \textbf{0.89} & \textbf{0.94} & 0.26 & 0.35 & 0.30 & 0.96 & \textbf{0.89} & 0.92 & 1.00 & 0.96 & 0.98\\
			glioma tumour & - & - & - & - & - & - & - & - & - & - & - & -\\
			meningioma tumour & \textbf{1.00} & \textbf{1.00} & \textbf{1.00} & \textbf{1.00} & \textbf{1.00} & \textbf{1.00} & \textbf{1.00} & \textbf{1.00} & \textbf{1.00}& 1.00 & 1.00 & 1.00\\
			metastasis tumour & \textbf{1.00} & \textbf{1.00} & \textbf{1.00} & 0.83 & 0.83 & 0.83 & 0.77 & 0.83 & 0.80 & 1.00 & 1.00 & 1.00\\
			tumour & \textbf{0.98} & \textbf{0.99} & \textbf{0.99} & 0.93 & 0.98 & 0.96 & 0.80 & 0.97 & 0.88 & 0.99 & 0.99 & 0.99\\
			subdural haematoma & \textbf{0.73} & \textbf{0.99} & \textbf{0.84} & 0.65 & 0.92 & 0.76 & 0.67 & 0.93 & 0.78 & 0.77 & 1.00 & 0.87\\
			small vessel disease & \textbf{0.93} & \textbf{0.98} & \textbf{0.95} & 0.57 & 0.85 & 0.69 & 0.91 & 0.89 & 0.90 & 0.95 & 0.97 & 0.96\\
			atrophy & \textbf{0.96} & \textbf{0.98} & \textbf{0.97} & 0.71 & 0.90 & 0.79 & 0.93 & 0.96 & 0.95 & 0.91 & .96 & .94\\
			microhaemorrhage & \textbf{0.90} & \textbf{0.90} & \textbf{0.90} & 0.00 & 0.00 & 0.00 & 0.83 & 0.50 & 0.63 & 1.00 & 1.00 & 1.00\\
			subarachnoid haemorrhage & 0.80 & \textbf{0.80} & 0.80 & 0.27 & 0.30 & 0.29 & \textbf{1.00} & \textbf{0.80} & \textbf{0.89} & 0.75 & 0.90 & 0.82\\
			haemorrhagic transformation & 0.25 & \textbf{1.00} & 0.40 & 0.00 & 0.00 & 0.00 & \textbf{1.00} & \textbf{1.00} & \textbf{1.00} & 0.50 & 1.00 & 0.67\\			
			location:cortical & \textbf{0.98} & \textbf{0.98} & \textbf{0.98} & 0.88 & 0.93 & 0.90 & 0.94 & 0.95 & 0.95 & 0.99 & 1.00 & 0.99\\
			location:deep & \textbf{0.92} & 0.88 & 0.90 & 0.90 & \textbf{0.96} & \textbf{0.93} & 0.90 & 0.88 & 0.90 & 0.95 & 0.94 & 0.94\\
			time:old & \textbf{0.98} & \textbf{0.97} & \textbf{0.97} & 0.94 & 0.97 & 0.95 & 0.92 & 0.90 & 0.91 & 0.97 & 0.98 & 0.98\\
			time:recent & \textbf{1.00} & \textbf{1.00} & \textbf{1.00} & 0.95 & 0.99 & 0.97 & 0.98 & 0.99 & 0.99& 1.00 & 1.00 & 1.00 \\
			\hline
			All & \textbf{0.94} & \textbf{0.96} & \textbf{0.95} & 0.82 & 0.92 & 0.86 & 0.91 & 0.94 & 0.92 & 0.96 & 0.98 & 0.97\\
		\end{tabular}
	\end{center}
	\caption{NER results and IAA scores on the ESS test data.}
	\label{tab:exp-ess-results}
\end{table}
\begin{table}[h!]
	\small
	\begin{center}
		\begin{tabular}{l|rrr|rrr|rrr||rrr}
			Evaluation on Tay test & \multicolumn{3}{c|}{EdIE-R} & \multicolumn{3}{c|}{EdIE-N}  & \multicolumn{3}{c}{SemEHR} & \multicolumn{3}{c}{IAA} \\
			\hline
			Entity Type  & P & R & F1 & P & R & F1 & P & R & F1 & P & R & F1\\
			 \hline		 
			ischaemic stroke & \textbf{1.00} & \textbf{0.99} & \textbf{1.00} & 0.78 & 0.97 & 0.86 & 0.99 & \textbf{0.99} & 0.99 & 0.90 & 1.00 & 0.96\\
			haemorrhagic stroke & \textbf{0.97} & 0.93 & \textbf{0.95} & 0.88 & \textbf{0.94} & 0.91 & 0.95 & 0.89 & 0.92 & 1.0 & 1.0 & 1.0\\
			stroke & \textbf{1.00} & \textbf{1.00} & \textbf{1.00} & 0.30 & 0.33 & 0.32 & 0.80 & 0.89 & 0.84 & - & - & - \\		
			glioma tumour & \textbf{0.80} & \textbf{0.44} & \textbf{0.57} & 0.00 & 0.00 & 0.00 & \textbf{0.80} & \textbf{0.44} & \textbf{0.57} & - & - & -\\
			meningioma tumour & \textbf{1.00} & \textbf{1.00} & \textbf{1.00} & \textbf{1.00} & \textbf{1.00} & \textbf{1.00} & \textbf{1.00} & \textbf{1.00} & \textbf{1.00} & - & - & -\\
			metastasis tumour & \textbf{1.00} & 0.99 & \textbf{1.00} & 0.88 & 0.94 & 0.91 & 0.91 & \textbf{1.00} & 0.95 & 1.00 & 1.00 & 1.00\\
			tumour & \textbf{0.99} & \textbf{1.00} & \textbf{0.99} & 0.88 & 0.95 & 0.91 & 0.98 & 0.59 & 0.74 & 0.89 & 1.00 & 0.94\\
			subdural haematoma & \textbf{0.92} & \textbf{1.00} & \textbf{0.96} & 0.70 & 0.89 & 0.78 & 0.82 & 0.79 & 0.80 & 1.00 & 1.00 & 1.00\\
			small vessel disease & 0.95 & \textbf{0.95} & \textbf{0.95} & 0.45 & 0.87 & 0.59 & 0.93 & 0.87 & 0.90 & 0.92 & 0.96 & 0.94\\
			atrophy & \textbf{1.00} & \textbf{0.97} & \textbf{0.99} & 0.81 & 0.93 & 0.86 & \textbf{1.00} & \textbf{0.97} & \textbf{0.99} & 1.00 & 1.00 & 1.00\\
			microhaemorrhage & \textbf{1.00} & \textbf{0.67} & \textbf{0.80} & 0.00 & 0.00 & 0.00 & 0.00 & 0.00 & 0.00 & 1.00 & 1.00 & 1.00 \\
			subarachnoid haemorrhage & \textbf{1.00} & 0.69 & 0.82 & 0.35 & 0.38 & 0.36 & \textbf{1.00} & \textbf{0.81} & \textbf{0.90} & 0.80 & 1.00 & 0.89\\
			haemorrhagic transformation & \textbf{1.00} & \textbf{1.00} & \textbf{1.00} & 0.00 & 0.00 & 0.00 & 0.00 & 0.00 & 0.00 & - & - & -\\
			location:cortical & \textbf{0.99} & \textbf{1.00} & \textbf{0.99} & 0.95 & 0.96 & 0.96 & 0.94 & 0.95 & 0.95 & 0.96 & 1.00 & 0.99 \\
			location:deep & \textbf{1.00} & \textbf{0.82} & \textbf{0.90} & 0.83 & 0.80 & 0.82 & 0.98 & 0.79 & 0.88 & 0.97 & 0.82 & 0.89\\
			time:old & \textbf{0.98} & \textbf{0.98} & \textbf{0.98} & 0.69 & 0.93 & 0.79 & 0.71 & 0.95 & 0.81 & 0.93 & 1.00 & 0.97\\
			time:recent & \textbf{0.99} & \textbf{0.99} & \textbf{0.99} & 0.92 & \textbf{0.99} & 0.95 & 0.98 & \textbf{0.99} & 0.98 & 0.98 & 1.00 & 0.99\\
			\hline
			All & \textbf{0.99} & \textbf{0.95} & \textbf{0.97} & 0.80 & 0.91 & 0.85 & 0.94 & 0.87 & 0.91 & 0.95 & 0.96 & 0.96 \\
		\end{tabular}
	\end{center}
	\caption{NER results and IAA scores on the Tay test dataset.}
	\label{tab:exp-tay-results}
\end{table}

\begin{table}[h!]
	\small
	\begin{center}
		\begin{tabular}{l|rrr|rrr|rrr}
			Evaluation on TayExt test & \multicolumn{3}{c|}{EdIE-R} & \multicolumn{3}{c|}{EdIE-N}  & \multicolumn{3}{c}{SemEHR} \\
			\hline
			Entity Type  & P & R & F1 & P & R & F1 & P & R & F1 \\
			 \hline		 
			ischaemic stroke & \textbf{0.98} & \textbf{0.93} & \textbf{0.95} & 0.75 & 0.91 & 0.82 & 0.94 & 0.85 & 0.89 \\
			haemorrhagic stroke & \textbf{0.86} & 0.74 & \textbf{0.79} & 0.68 & 0.64 & 0.66 & 0.81 & \textbf{0.75} & 0.78 \\
			stroke & \textbf{1.00} & 0.60 & 0.75 & 0.25 & 0.40 & 0.31 & 0.80 & \textbf{0.80} & \textbf{0.80} \\
			glioma tumour & \textbf{0.86} & \textbf{1.00} & \textbf{0.92} & 0.00 & 0.00 & 0.00 & 0.80 & 0.33 & 0.47 \\
			meningioma tumour & \textbf{1.00} & \textbf{1.00} & \textbf{1.00} & \textbf{1.00} & \textbf{1.00} & \textbf{1.00} & \textbf{1.00} & \textbf{1.00} & \textbf{1.00} \\					
			metastasis tumour & \textbf{1.00} & \textbf{1.00} & \textbf{1.00} & 0.83 & 0.86 & 0.85 & \textbf{1.00} & \textbf{1.00} & \textbf{1.00} \\								tumour & \textbf{0.95} & 0.86 & \textbf{0.91} & 0.73 & \textbf{0.92} & 0.82 & 0.71 & 0.76 & 0.74 \\		
			subdural haematoma & \textbf{0.86} & \textbf{0.87} & \textbf{0.86} & 0.61 & 0.73 & 0.66 & 0.76 & 0.85 & 0.78 \\
			small vessel disease & \textbf{0.95} & \textbf{0.84} & \textbf{0.89} & 0.36 & 0.72 & 0.48 & 0.89 & 0.75 & 0.82 \\				
			atrophy & \textbf{1.00} & 0.94 & 0.97 & 0.89 & \textbf{0.94} & 0.92 & 0.98 & \textbf{0.98} &  \textbf{0.98} \\
			microhaemorrhage & \textbf{1.00} & \textbf{0.50} & \textbf{0.67} & 0.00 & 0.00 & 0.00 & 0.00 & 0.00 & 0.00 \\
			subarachnoid haemorrhage & 0.94 & 0.63 & 0.76 & 0.36 & 0.35 & 0.36 & \textbf{0.96} & \textbf{0.83} & \textbf{0.89} \\
			haemorrhagic transformation & 0.28 & \textbf{0.70} & \textbf{0.40} & 0.00 & 0.00 & 0.00 & \textbf{0.33} & 0.50 & \textbf{0.40} \\
			location:cortical &\textbf{ 0.98} & \textbf{0.99} & \textbf{0.99} & 0.94 & 0.97 & 0.95 & 0.96 & 0.98 & 0.97 \\
			location:deep & 0.91 & \textbf{0.96} & 0.93 & 0.77 & 0.87 & 0.81 & \textbf{0.95} & 0.93 & \textbf{0.94} \\
			time:old & \textbf{1.00} & \textbf{0.93} & \textbf{0.96} & 0.78 & 0.92 & 0.84 & 0.86 & 0.89 & 0.87 \\
			time:recent & \textbf{1.00} & 0.95 & \textbf{0.97} & 0.90 & 0.93 & 0.92 & 0.94 & \textbf{0.97} & 0.96 \\
			\hline
			All & \textbf{0.94} & \textbf{0.91} & \textbf{0.93} & 0.76 & 0.85 & 0.80 & 0.89 & 0.88 & 0.89 \\
		\end{tabular}
	\end{center}
	\caption{NER results on the TayExt test dataset.}
	\label{tab:exp-tayext-results}
\end{table}